\documentclass{interact}

\usepackage{biblatex}
\usepackage{caption}
\usepackage{subcaption}
\usepackage{multirow}
\usepackage{url}

\begin{document}


\title{Development and external validation of a lung cancer risk estimation tool using gradient-boosting}

\author{
\name{Pierre-Louis Benveniste\textsuperscript{1*}, Julie Alberge\textsuperscript{2,1*}, Lei Xing\textsuperscript{3}, Jean-Emmanuel Bibault\textsuperscript{4,5}}
\affil{\textsuperscript{*} contributed equally}
\affil{\textsuperscript{1} Ecole Normale Supérieure de Paris Saclay}
\affil{\textsuperscript{2} École Nationale des Ponts et Chaussées}
\affil{\textsuperscript{3} Laboratory for Artificial Intelligence in Medicine, Stanford University}
\affil{\textsuperscript{4} Radiation Oncology Department, Hôpital Européen Georges Pompidou, AP-HP, Université Paris Cité}
\affil{\textsuperscript{5} HeKA Team, Inria Inserm UMR 1138, PariSantéCampus, Paris, France}
\affil{\textbf{\\Corresponding author:}}
\affil{Jean-Emmanuel Bibault}
\affil{Radiation Oncology Department, Hôpital Européen Georges Pompidou, AP-HP}
\affil{20 rue Leblanc, 75015, Paris, France}
\affil{jean-emmanuel.bibault@aphp.fr}
}

\maketitle

\begin{abstract}
\indent Introduction: Lung cancer is a significant cause of mortality worldwide, emphasizing the importance of early detection for improved survival rates. In this study, we propose a machine learning (ML) tool trained on data from the PLCO Cancer Screening Trial and validated on the NLST to estimate the likelihood of lung cancer occurrence within five years.

Methods: The study utilized two datasets, the PLCO (n=55,161) and NLST (n=48,595), consisting of comprehensive information on risk factors, clinical measurements, and outcomes related to lung cancer. Data preprocessing involved removing patients who were not current or former smokers and those who had died of causes unrelated to lung cancer. Additionally, a focus was placed on mitigating bias caused by censored data. Feature selection, hyper-parameter optimization, and model calibration were performed using XGBoost, an ensemble learning algorithm that combines gradient boosting and decision trees.

Results: The final ML model was trained on the pre-processed PLCO dataset and tested on the NLST dataset. The model incorporated features such as age, gender, smoking history, medical diagnoses, and family history of lung cancer. The model was well-calibrated (Brier score=0.044). ROC-AUC was 82\% on the PLCO dataset and 70\% on the NLST dataset. PR-AUC was 29\% and 11\% respectively. When compared to the USPSTF guidelines for lung cancer screening, our model provided the same recall with a precision of 13.1\% vs. 9.3\% on the PLCO dataset and 3.2\% vs. 3.1\% on the NLST dataset.

Conclusion: The developed ML tool provides a freely available web application for estimating the likelihood of developing lung cancer within five years. By utilizing risk factors and clinical data, individuals can assess their risk and make informed decisions regarding lung cancer screening. This research contributes to the efforts in early detection and prevention strategies, aiming to reduce lung cancer-related mortality rates.
\end{abstract}


\begin{github_repo}
https://github.com/plbenveniste/LungCancerRisk
\end{github_repo}

\section{Introduction}
Cancer is a leading cause of death worldwide, accounting for nearly 10 million deaths in 2020, or nearly one in six deaths \cite{cancer_site}. Lung cancer is the most common cause of cancer death in 2020 with around 1.80 million deaths. The survival rate of lung cancer is strongly dependent on the cancer stage as well as the physical condition of the patient. On average, it is estimated that the five-year survival rate for lung cancer is around 56\% for cases detected when the disease is still localized within the lungs. On the other hand, in later stages, when the disease has spread to other organs, the five-year survival rate drops to 5\% \cite{American_Lung_Association_undated-ha}, highlighting the need for early detection. In response, different recommendations have been made. Based on the National Lung Screening Trial (NLST) \cite{nlst}, the United States Preventive Services Task Force recommends lung cancer screening with low-dose computed tomography (LDCT) in adults aged 55 to 80 years who have a 30 pack-year smoking history and are currently smoking or have quit within the past 15 years \cite{Moyer2014-va}. The conclusion of the NLST was that screening using low-dose computed tomography (LDCT) resulted in a decrease in mortality equal to 3 fewer deaths per 1,000 participants \cite{Tanoue2015-xq}. This study amongst others such as DANTE (Detection and Screening of Early Lung Cancer by Novel Imaging Technology and Molecular Essays) or the DLCST (Danish Lung Cancer Screening Trial) also studied different strategies for screening, the harms and radiations caused by screening and other factors related to lung cancer mortality. Known risk factors are key indicators when identifying patients with high risks of lung cancer occurrence \cite{Malhotra2016-xi}. In this study, we chose to focus on patients who are current or former smokers.

We propose a machine learning (ML) tool to compute the likelihood of lung cancer occurrence trained on data from the Prostate, Lung, Colorectal, and Ovarian (PLCO) Cancer Screening Trial and validated on the National Lung Screening Trial (NLST). From this ML-based tool, we developed a freely available web application that people can use to estimate their likelihood of developing lung cancer and sensitize them to lung cancer screening for early detection of lung cancer.

\section{Data used}

\subsection{Data sources}

Data used in this project come from two different datasets. The Prostate, Lung, Colorectal, and Ovarian (PLCO) Cancer Screening Trial \cite{plco} was used as the training dataset. It contains data from 154,887 patients with 219 features for each. The PLCO Screening Trial is a significant study evaluating cancer screening tests for prostate, lung, colorectal, and ovarian cancers. It collects comprehensive data on risk factors, clinical measurements, and outcomes. The trial assessed screening effectiveness and identified potential risk factors for specific cancers. Furthermore, the second dataset used is the National Lung Screening Trial (NLST) \cite{nlst}. It was used as the external testing dataset. Conducted by the National Cancer Institute (NCI), the NLST study aimed to evaluate the efficacy of low-dose computed tomography (LDCT) \cite{Tammemagi2014-mk} in detecting lung cancer among individuals at high risk. This extensive dataset includes comprehensive information, ranging from demographic characteristics and smoking history to imaging findings and participant outcomes, encompassing over 53,452 individuals with 324 features each.

\subsection{Risk factors}

In order to identify patients with a high risk of developing lung cancer, we target those who meet certain criteria. Those criteria are some of the risk factors related to lung cancer. The primary risk factor, and one that is responsible for the majority of cases, is tobacco smoking. The carcinogens present in tobacco smoke can damage the DNA of lung cells, leading to the initiation and progression of cancer. Other modifiable risk factors include exposure to secondhand smoke, occupational exposure to carcinogens such as asbestos and radon, and air pollution. Non-modifiable factors, such as a family history of lung cancer and certain genetic mutations, also play a role in increasing susceptibility. Additionally, factors such as age, gender, and a history of lung diseases like chronic obstructive pulmonary disease (COPD) further contribute to the risk profile. Understanding and addressing these multifaceted risks are crucial for effective prevention and early detection strategies in combating lung cancer \cite{Malhotra2016-xi}. We used only former or current smokers and decided to remove participants who have never smoked from the PLCO dataset leaving 80,668 participants in the dataset (52\% of the original dataset). This is even more relevant as the NLST dataset only studies former and current smokers.

\subsection{Feature extraction}

Based on the previous risk factors, we will describe the features extracted from both PLCO and NLST. These features have been pre-processed so that both datasets match in terms of column names, data type, and formatting. All the attributes of the participants are described in Appendix \ref{app:list_feature}. The final model doesn’t require as many as we will show in the next sections that some of these features play a minor role and barely modify the precision and recall of the model. 

\subsection{Data pre-processing}

In this section, we describe the pre-processing of both the training and testing datasets, respectively PLCO and NLST. The first step involved removing patients who are not current or former smokers since NLST only contains current or former smokers. This pre-processing step removed 74,219 patients from the PLCO dataset (around 48\%), leaving 80,668  patients. Then, we started by removing patients who had died of something else than lung cancer. By removing these patients we removed a bias in favor of lung cancer occurrence. Indeed, as some of these patients were suffering from severe medical conditions, their weak physical condition favored the development of tumors. This pre-processing removed 24,356 patients from the PLCO dataset (around 16\%), leaving 56,312 patients as well as removed 2,904 patients from the NLST dataset (around 5\%), leaving 50,548 patients. As a third step, we focused on removing bias brought by censored data \cite{Gijbels2010-wc}. Both in NLST and PLCO, the duration of a patient’s study varies widely from one participant to another introducing bias into the model. Indeed, a participant who was studied for less than a year has less chance of being diagnosed with lung cancer than a participant who remained for 7 years. On the other hand, we decided to keep the participants who stayed longer than average in the study: the reason is that, even though they introduce a bias in favor of positive screening for lung cancer, we would rather have more false positives than false negatives (favoring recall over precision). We decided for the sake of predicting lung cancer risk in the next 5 years to train the model on patients from PLCO who were either negative for lung cancer screening and studied for longer than 2100 days (5.75 years) or positive for lung cancer screening. This preprocessing step removed 1,151 patients from PLCO (around 1\%), leaving 55,161 patients as well as removed 1,953 patients from NLST (around 4\%), leaving 48,595 patients. 

\section{Methods}

The development of this model required feature selection, hyperparameter optimization, and calibration of the model.

\subsection{XGBoost}

We trained an XGBoost model \cite{Chen2016-ls} on the pre-processed PLCO dataset and tested it on NLST. XGBoost combines the principles of gradient boosting \cite{Friedman2002-bs} and decision trees \cite{deville} to create a robust and high-performance predictive model. The algorithm iteratively builds an ensemble of weak prediction models, called decision trees, by optimizing an objective function that measures the model's performance. During each iteration, XGBoost computes the gradient of the loss function and updates the tree ensemble by adding new trees that minimize the residual errors. One of the key strengths of XGBoost lies in its ability to handle a wide range of data types and feature interactions effectively. It incorporates several advanced techniques, such as regularization, parallel processing, and column blocking, to optimize both accuracy and computational efficiency. XGBoost can automatically handle missing values and can capture non-linear relationships, making it highly versatile. Moreover, XGBoost provides various hyperparameters that allow researchers to fine-tune the model for their specific needs. These hyperparameters control the depth, learning rate, number of trees, and regularization, among others, enabling customization and model optimization.
XGBoost was chosen for this project as it can deal with missing values which are numerous in both datasets. The goal of the XGBoost in our case was to output 1 if the person is likely to develop lung cancer in the next 5 years and 0 otherwise. 

\subsection{Feature selection}

Since our final objective was to build a self-evaluation tool to promote lung cancer screening for people at risk, we wanted to design a tool that relies on few features. After exploring different combinations of features, we selected the most important features using Shapley values \cite{Lundberg2017-xj}.

\subsection{Hyperparameter optimization}

To improve the results of the XGBoost model we perform hyperparameter optimization. The methodology used for hyperparameter optimization relies on Bayesian Grid Search \cite{Bergstra_undated-kj}. Bayesian optimization is a powerful technique employed in the field of machine learning to efficiently search and tune hyperparameters for predictive models. When applied to XGBoost, Bayesian optimization enables the automatic optimization of its hyperparameters, taking into account the AUC-PR metric \cite{Sofaer2019-bs} for evaluating model performance. The AUC-PR metric is particularly relevant for imbalanced classification problems, such as lung cancer prediction, where accurate identification of positive cases is crucial. By incorporating Bayesian optimization with the AUC-PR metric, we leveraged the strengths of XGBoost and enhanced its predictive capabilities for lung cancer prediction. The iterative nature of Bayesian optimization intelligently explores the hyperparameter space, directing the search toward promising regions and gradually refining the model's configuration. 
During Bayesian optimization we focused on the following parameters:
\begin{itemize}
    \item Learning rate: The rate at which the model learns from the data during training.
    \item Max depth: The maximum depth or levels of the decision tree or random forest model.
    \item Subsample: The fraction of samples used for training each individual tree in the ensemble.
    \item Colsample\_bytree: The fraction of features or columns used for training each individual tree.
    \item N estimators: The number of trees or estimators in the ensemble model.
\end{itemize}

\subsection{Calibration}

Plotting a calibration curve serves as a valuable tool for assessing the reliability and accuracy of the model's predictions. A calibration curve visually compares the predicted probabilities or scores generated by the model with the observed outcomes or true probabilities. This graphical representation plays a crucial role in evaluating model calibration, which measures the agreement between predicted probabilities and observed frequencies of the target outcome. A well-calibrated model should exhibit predictions that closely align with the actual probabilities, as indicated by a calibration curve closely following the diagonal line. Additionally, the calibration curve allows for the measurement of confidence in the model's predictions, enabling researchers to determine if the predicted probabilities accurately reflect the likelihood of the target outcome. Deviations from the diagonal line may indicate overconfidence or underconfidence in the model's estimates. Consequently, the calibration curve aids in selecting an appropriate decision threshold for classification tasks by identifying the range of predicted probabilities where the model is well-calibrated. Overall, the calibration curve serves as a critical tool for evaluating model performance and guiding adjustments to enhance its reliability and accuracy.

\subsection{Interpretability}

By assigning importance values to input features, Shapley values \cite{Lundberg2017-xj} offer an intuitive approach to understanding model predictions. In the context of lung cancer prediction, Shapley values provide valuable insights into the individual feature importance, allowing patients to identify the most influential factors contributing to the predicted risk of lung cancer. Moreover, Shapley values enable both global and local interpretability, shedding light on the overall impact of features across the dataset and providing case-specific explanations for individual predictions. Considering the complexity of XGBoost models and their ability to capture feature interactions, Shapley values offer an advantageous perspective by accounting for joint contributions of features. This comprehensive understanding of feature interactions enhances the interpretability of the model and contributes to a more nuanced comprehension of how different combinations of features influence the prediction of lung cancer risk. Additionally, Shapley values can aid in assessing model fairness and bias, enabling the detection of disparities in feature importance across different subgroups. Such information is essential for ensuring the fairness and equity of healthcare decision-making processes. Ultimately, the utilization of Shapley values for interpretability in the domain of lung cancer prediction using XGBoost not only improves model transparency and performance but also supports the development of reliable and trustworthy prediction models in healthcare settings. 

\section{Results}

After data pre-processing, the PLCO dataset contains 55,161 patients and the NLST dataset contains 48,595 patients. In this section, we describe both datasets and the distribution of patients for each feature (Table \ref{tab:feature_descr} in Appendix). 

\subsection{Model performance}

The XGBoost model is trained on the PLCO dataset with the following parameters. Firstly, there are two types of booster parameters: linear models and tree-based models. In this study we used a tree-based model as it makes sense in the case of a binary classification and because it usually outperforms the linear models. For the loss function, we used a binary logistic objective function. This loss function is designed as a logistic regression for binary classification (which is our case) and returns a predicted probability here corresponding to the likelihood of the person developing lung cancer in the next 5 years. Furthermore, we chose to use an exact tree method because it is more precise and the training in our project is relatively short. Finally, for the evaluation metric we chose to use the Area Under the Precision Recall Curve (AUC-PR).
Before diving into the hyperparameter optimization, we train a basic XGBoost on the training set. Using the Shap library \cite{Lundberg2017-xj}, we look at the most important features and select the most contributing ones. Selecting these features is important for us as our goal is to design a tool on which people can easily evaluate their risk of developing lung cancer. Furthermore, we know that extensive and long questionnaires play a significant role in barriers to entry. We chose to simplify our model and slightly downgrade the overall results of the model to retain more patients. The following Figure \ref{fig:shap} details the most contributing features. The following features were selected:
\begin{itemize}
    \item Age
    \item Smoking cessation age
    \item Cigarette smoking
    \item Smoking onset age
    \item Cigarette per day
    \item Pack years
    \item Smoke years
    \item Lung cancer family history
    \item BMI
\end{itemize}

\begin{figure}[h]
    \centering
    \captionsetup{justification=centering}
    \includegraphics[scale=0.5]{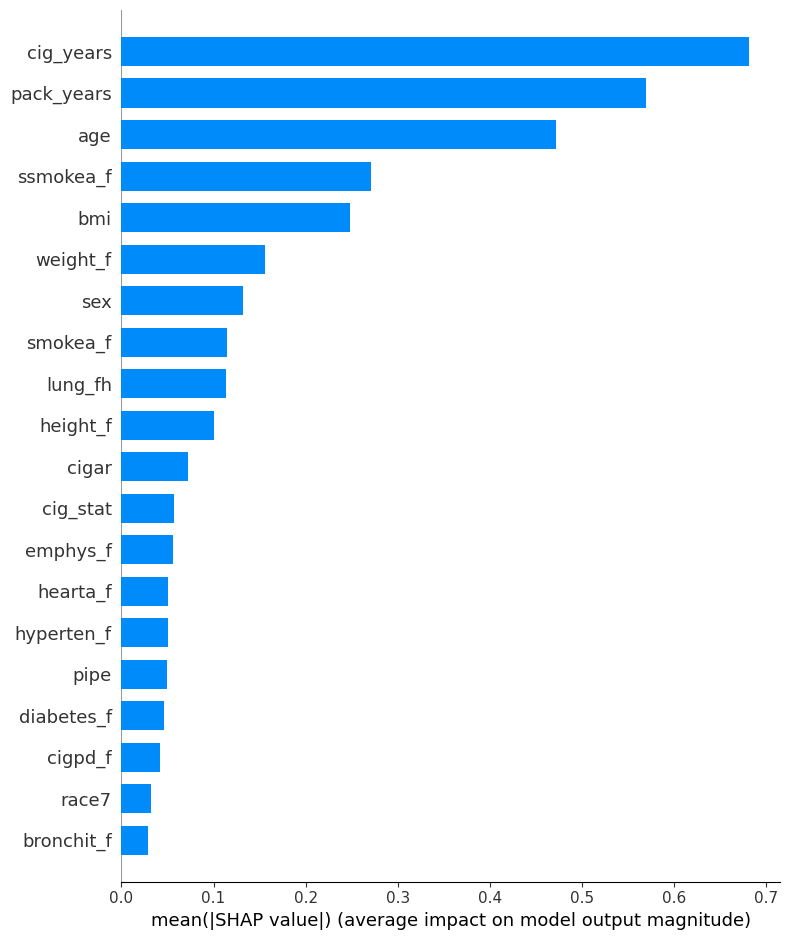}
    \caption{Shapley values of the features used for lung cancer risk prediction}
    \label{fig:shap}
\end{figure}

\begin{figure}[h!]
    \centering
    \captionsetup{justification=centering}
    \includegraphics[scale=0.5]{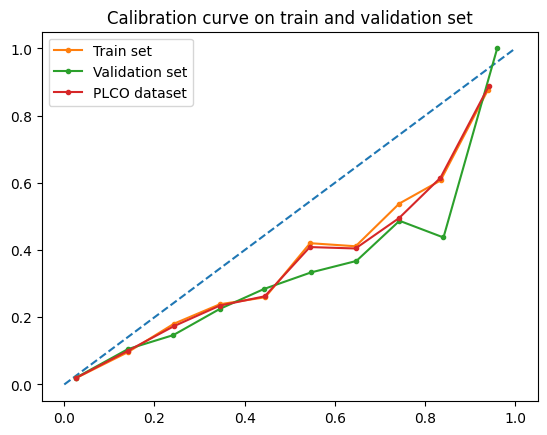}
    \caption{Calibration of the model on the PLCO and NLST datasets}
    \label{fig:calib}
\end{figure}

After hyperparameter optimization, ROC AUC was 0.83 on the PLCO validation dataset and 0.69 on the NLST external testing dataset (Appendix \ref{app:roc}). The results obtained by our model show a good performance on the validation set (Table \ref{tab:model_perf}). While the model calibration was good (Figure \ref{fig:calib}), the curve suggests under-confidence in the predictions of the model while still performing relatively well as the results lie close to the y=x line (in blue). However, when focusing on calibration-in-the-small, it seems that the model is underconfident for people with a high risk of developing lung cancer. This shows the importance of recall in this case. Indeed because, the model tends to under-classify some of the high-risk patients, focusing on having a high recall is key, even though it might lower the precision of the model. To do so, one must select a lower classification threshold to be less selective when predicting lung cancer. 

\begin{table}[h]
    \centering
    \captionsetup{justification=centering}
    \begin{tabular}{|l|l|l|}
        \hline
        \textbf{Metric} & \textbf{Validation Dataset} & \textbf{Testing Dataset} \\
        \hline
        ROC AUC & 0.82 & 0.70 \\
        Brier Score & 0.043 & 0.044 \\
        Average Precision & 0.52 & 0.14 \\
        Average Recall & 0.02 & 0.04 \\
        AUC-PR & 0.29 & 0.11 \\
        \hline
    \end{tabular}
    \caption{Model performances on the PLCO and NLST datasets}
    \label{tab:model_perf}
\end{table}

\subsection{Comparison with USPSTF recommendations}

\begin{figure}[h!]
    \centering
    \captionsetup{justification=centering}
    \includegraphics[width=\textwidth]{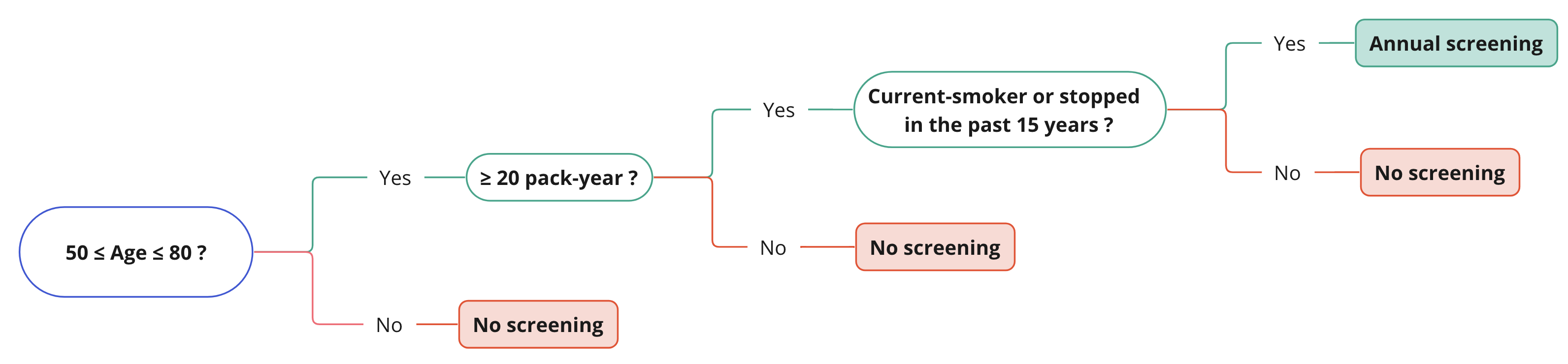}
    \caption{USPSTF decision tree}
    \label{fig:decision_tree}
\end{figure}

The US Preventive Services Task Force issued a recommendation statement in 2021 for lung cancer screening. This was an update from the 2013 recommendation. These recommendations can be summarized as a decision tree (Figure \ref{fig:decision_tree}). This tool was designed based on NLST. Out of the 48,595 patients who are described in the NLST dataset, 48,034 of them fit into the criteria made by the USPSTF recommendation statement. It makes sense that a lot of them fit into these criteria since they were built using this dataset. Among them, 1,495 of them had cancer while overall in the dataset 1,511 had cancer. This means that overall in the NLST dataset, the criteria has a recall of around: 98.9\%. Of the 55,161 patients who participated in the PLCO study (and who remained after pre-processing), 22,609 of them fit into the criteria of the USPSTF recommendation statement. Among those who fit into the criteria of the USPSTF recommendation statement, 2,105 had cancer while overall in the whole dataset, 2,752 had cancer. This means that overall PLCO, the criteria has a recall of around 76.5\% in identifying the patients that have cancer. It is interesting to see that, while their criteria work very well on the NLST dataset, which makes sense because they used it to build it, it doesn't perform as well on the PLCO dataset (Table \ref{tab:uspstf}).

\begin{table}[h]
    \captionsetup{justification=centering}
    \centering
    \begin{tabular}{|l|l|l|}
    \hline
    \textbf{Metric} & \textbf{NLST} & \textbf{PLCO} \\
    \hline
    Total number of participants* & 48,595 & 55,161 \\
    TP & 1,495 & 2,105 \\
    FN & 16 & 647 \\
    FP & 46,539 & 31,905 \\
    TN & 545 & 20,504 \\
    Precision & 3.1\% & 9.3\% \\
    Recall & 98.9\% & 76.5\% \\
    \hline
    \end{tabular}
    \caption{USPSTF recommendations results on the pre-processed dataset (*after pre-processing)}
    \label{tab:uspstf}
\end{table}

Moreover, we can compare the precision and recall of the US Recommendation tool with our model. In the following table (Table \ref{tab:recall_preci_our}), by fixing equal recall, we can observe the precision level. The improvement is small for the NLST and we can explain that by the fact that the US Recommendation tool was designed based on NLST. However, we can see a clear impact on the PLCO recall (Table \ref{tab:recall_preci_our}).

\begin{table}[h]
    \captionsetup{justification=centering}
    \centering
    \begin{tabular}{|l|l|l|l|}
    \hline
     \multicolumn{2}{|l|}{} & \textbf{Recall} & \textbf{Precision} \\
    \hline
    \multirow{2}{*}{\textbf{NLST}} & USPSTF Recommendation & 98.9\% & 3.1\% \\
    & Our Model & 98.9\% &\textbf{3.2\%} \\
    \hline
    \multirow{2}{*}{\textbf{PLCO}} & USPSTF Recommendation & 76.5\% & 9.3\% \\
    & Our Model & 76.5\% & \textbf{13.1\%} \\
    \hline
    \end{tabular}
    \caption{Recall and Precision of USPSTF and our model}
    \label{tab:recall_preci_our}
\end{table}

\subsection{Online prediction tool}

As part of this project, we also built a web application \cite{web_app}. The goal of this app is to be accessible to all for people to assess the risk of them developing cancer in the next 5 years. The risk prediction is based on a short 8-question form, which are the features used in the model. We used the model saved as a pickle file and Python files hosted on GitHub \cite{lungcancerapp}. Heroku \cite{Middleton2013-he} deploys and maintains the app: it's a Platform as a Service (PaaS) tool. The web application is currently online and working. 

\section{Discussion}

A data-driven risk model was established and tested to predict the 5-year outcomes related to National Lung Screening Trial (NLST)-like CT lung cancer screenings. This model demonstrated robust validation within U.S. research groups (PLCO and NLST), indicating their wide applicability. The main finding from our model suggests that selecting smokers based on individual risk, as opposed to risk-factor-based groups, could potentially prevent more deaths, enhance screening effectiveness, and increase screening efficiency. 

Several models have been created to better identify smokers who should be screened for lung cancer Katki et al developed and validated risk models for lung cancer screening using low-dose CT \cite{Katki2016-vc}. The model used factors like age, race, smoking history, and family history. Screening based on these models was more effective than USPSTF guidelines in preventing lung cancer deaths. The model identified high-risk individuals not eligible under USPSTF criteria, including current and former smokers. These risk-based screenings have been shown to potentially prevent 90\% of CT-preventable lung cancer deaths by screening only 49\% of American smokers aged 50 to 80 \cite{Friedman2002-bs}. These methods outperform USPSTF recommendations by favoring high-risk, high-benefit smokers who might not be eligible under USPSTF guidelines, but who have a higher 5-year lung cancer risk and a lower number needed to screen (NNS). These high-risk individuals cannot be identified by groups and necessitate risk calculations. Meanwhile, a substantial proportion of USPSTF-eligible individuals are at a lower risk, meaning they might not benefit as much from screening without a risk calculation. Risk-based selection could also increase the number of African Americans and women chosen for CT lung screening.

In 2022, Kumar et al used machine learning and deep learning techniques to predict the growth and progression of lung cancer \cite{Anil_Kumar2022-bl}. They built prediction models using supervised machine learning algorithms and analyzed images using the local binary patterns (LBP) technique. This study proposed a machine learning model based on support vector machines (SVM) for the detection of lung cancer using symptom classification. Data acquisition and preprocessing were performed on a dataset from the University of California, Irvine. The same year, Dritsas et al used a public dataset \cite{Bhat2021-ex} containing 309 patients with 15 features each and employed several techniques to improve the accuracy of the models \cite{Dritsas2022-jl}. These techniques included class balancing, which addressed the imbalance in the dataset, and feature ranking, which determined the most important features for prediction. Among the different classification models tested, the Rotation Forest model demonstrated the highest efficiency in predicting lung cancer occurrence. 

In the context of lung cancer prediction using machine learning, several performance metrics are commonly employed to evaluate the effectiveness of the model and understand its predictive capabilities. When evaluating the performance of a lung cancer prediction model, the choice of the performance metric is crucial in capturing the specific characteristics of the problem at hand. While metrics like ROC AUC are commonly used in the mentioned studies, and provide a comprehensive evaluation of a model's performance across different classification thresholds, Area-Under Precision-Recall Curve (AUC-PR) offers distinct advantages in certain scenarios, such as imbalanced datasets or when the focus is on positive instances. This metric considers the trade-off between precision and recall across different classification thresholds, providing a holistic measure of the model's performance. In lung cancer screening, there is a high population imbalance, i.e. the number of negative instances (non-cancerous cases) is significantly higher than the number of positive instances (cancerous cases). In such cases, using ROC AUC as the sole performance metric can be misleading because it emphasizes the model's ability to rank all instances correctly, irrespective of the class distribution. AUC-PR, on the other hand, places more emphasis on correctly predicting positive instances, which is crucial in the context of lung cancer prediction. By prioritizing positive predictions, AUC-PR provides a more meaningful evaluation of the model's performance in identifying lung cancer cases. Furthermore, lung cancer prediction is often a critical task where the goal is to minimize false negatives (missed diagnoses) while maintaining a reasonable level of precision. In our study,  AUC-PR was 29\% on the PLCO dataset and 11\% on the NLST. Our model has the same Precision as the USPSTF guidelines, with better Recall (9.3\% vs. 13.1\% on PLCO and 3.1\% vs. 3.2\% on NLST).

There are significant limits to risk-based eligibility for lung screening. Models tend to favor the elderly, which can lead to saving fewer life-years and decrease cost-efficiency. These models can also be biased towards selecting patients with comorbidities such as COPD. The next generation of models could be trained to predict life-years saved from screening. Our study has some significant drawbacks: the training and testing data were collected in the US only and we can't assume that our findings can be generalized beyond the US. Even if the data was collected from two prospective trials, there were still missing values. Also, the two trials were initially not designed to answer the same question with the same procedure, which could potentially bias the results. There is a trade-off between creating the best model for defining screening eligibility and creating a model that can be implemented effectively in the daily routine. Some risk models have web-based tools, like the Risk-based NLST Outcomes Tool for LCDRAT and LCRAT \cite{rnot}, and MyLungRisk for LLPv2 \cite{mylungrisk}. These models do not provide the most important features involved in the prediction, but our interface does.  Getting an accurate prediction is of course interesting from the clinical point of view, but getting a prediction along with the most important feature means that the prediction can become actionable. Finally,  these tools could be integrated into electronic medical software to automatically calculate the risk of any individual.

\section{Conclusion}

We created a model to predict lung cancer risk at five years using XGBoost with better precision and recall than the current USPSTF recommendations. Implementing risk-based screening in clinical practice can be challenging and requires accurate, user-friendly decision aids to support shared decision-making. The web risk calculator can be used to directly communicate personalized risk prediction to the patient. In that context, shared decision-making processes should be carefully evaluated, as most lung cancer deaths are currently not preventable through screening, even if CT screening can reduce lung cancer mortality by 20\%.

\printbibliography

\appendix

\section{List of features}\label{app:list_feature}
\begin{itemize}
    \item Age: This feature captures the person’s age.
    \item Gender: This feature shows if the person is male or female.
    \item Height: This feature indicates the person’s height in inches.
    \item Weight: This feature indicates the person’s weight in pounds.
    \item Race: This feature describes which race the person is among the following (White or Hispanic, Black, Asian, Pacific Islander, American Indian, missing).
    \item Smoking cessation age: This feature describes the age at which the person stopped smoking.
    \item Cigarette smoking: This feature describes if the person is a current or a former cigarette smoker at the beginning of the study.
    \item Cigar smoking: This feature describes if the person is a current or a former cigar smoker at the beginning of the study.
    \item Pipe smoking: This feature describes if the person is a current or a former pipe smoker at the beginning of the study.
    \item Pack years: This feature refers to the number of packs smoked per day multiplied by the number of years during which the person smoked.
    \item Smoking onset age: This feature indicates the age at which the person started smoking.
    \item Cigarette per day: This feature refers to the number of cigarettes smoked per day.
    \item Smoke years: This feature describes the total number of years during which the person smoked.
    \item Age chronic bronchitis: This feature describes the age at which the person was diagnosed with chronic bronchitis.
    \item Diabetes diagnosis: This feature describes if the person was ever diagnosed with diabetes.
    \item Emphysema diagnosis: This feature indicates if the person was ever diagnosed with emphysema.
    \item Heart diagnosis: This feature indicates if the person was ever diagnosed with a heart disease or a heart attack.
    \item Hypertension diagnosis: This feature indicates if the person was ever diagnosed with hypertension.
    \item Stroke diagnosis: This feature indicates if the person was ever diagnosed with a stroke.
    \item BMI: This feature describes the person’s body mass index.
    \item Lung cancer family history: This feature describes if the person has close family (parents, siblings, or child) who had lung cancer.
    \item Lung cancer: This feature indicates if the person was diagnosed with lung cancer.
\end{itemize}

\section{ROC-AUC curve}\label{app:roc}

\begin{figure}[h!]
    \centering
    \captionsetup{justification=centering}
    \includegraphics[scale=0.6]{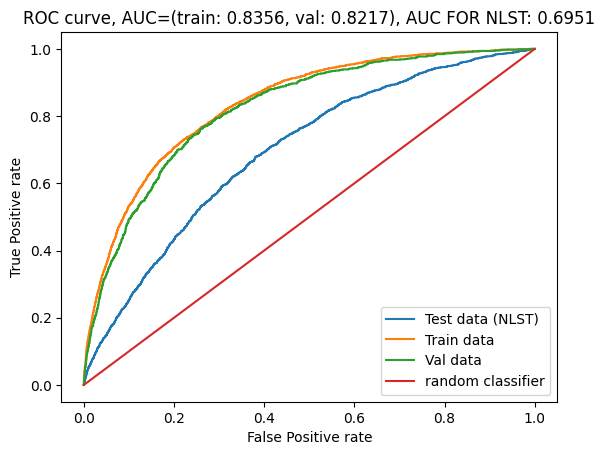}
    \caption{ROC-AUC curve of the model on the train, validation (PLCO), and external test dataset (NLST)}
    \label{fig:roc}
\end{figure}
\newpage
\section{Characteristics and statistics for PLCO and NLST groups}

\begin{table}[h!]
\centering
\captionsetup{justification=centering}
\begin{tabular}{|c|l|l|l|}
\hline
\multicolumn{2}{|l|}{\multirow{3}{*}{\textbf{Characteristics}}} & \multicolumn{2}{|l|}{\textbf{No (\%)}} \\
\cline{3-4}
\multicolumn{2}{|l|}{} & \textbf{PLCO} & \textbf{NLST} \\
\cline{3-4}
\multicolumn{2}{|l|}{} & 55,161 & 48,595 \\
\hline
\multirow{4}{*}{\textbf{Age}} & $\leq 50$ & 0 (0.0) & 1 (0.0) \\
& 51-60 & 27,337 (49.6) & 24,861 (51.2) \\
& 61-70 & 25,120 (45.5) & 20,901 (43.0) \\
& $> 70$ & 2,704 (4.9) & 2,832 (5.8) \\
\hline
& $\leq 30$ & 10,470 (19.0) & 2 (0.0) \\
& 31-40 & 11,886 (21.5) & 130 (0.3) \\
\textbf{Smoking cessation}& 41-50 & 11,447 (20.8) & 7,025 (14.5) \\
\textbf{age}& 51-60 & 8,649 (15.7) & 14,071 (29.0) \\
& $> 60$ & 1,942 (3.5) & 4,378 (9.0) \\
& Missing & 10,767 (19.5) & 22,989 (47.3) \\
\hline
\multirow{3}{*}{\textbf{Smoking Status}} & Active smoker & 9,965 (18.1) & 22,842 (47.0) \\
& Former smoker & 45,196 (81.9) & 25,753 (53.0) \\
& Missing & - & - \\
\hline
\multirow{4}{*}{\textbf{Pack-years}} & $\leq 25$ & 26,981 (48.9) & 8 (0.0) \\
& 26-50 & 16,147 (29.3) & 26,746 (55.0) \\
& 51-100 & 9,448 (17.1) & 19,544 (40.2) \\
& $> 100$ & 1,434 (2.6) & 2,297 (4.7) \\
& Missing & 1,151 (2.1) & 0 (0.0) \\
\hline
\multirow{4}{*}{\textbf{Smoking onset age}} & $\leq 15$ & 10,169 (18.4) & 17,927 (36.9) \\
& 16-20 & 33,760 (61.2) & 25,411 (52.3) \\
& $> 30$ & 10,950 (19.9) & 5,256 (10.8) \\
& Missing & 282 (0.5) & 1 (0.0) \\
\hline
\multirow{6}{*}{\textbf{Smoking-years}} & $\leq 10$ & 8,800 (16.0) & 2 (0.0) \\
& 11-20 & 11,761 (21.3) & 292 (0.6) \\
& 21-30 & 11,532 (20.9) & 5,134 (10.6) \\
& 31-40 & 13,037 (23.6) & 21,620 (44.5) \\
& $> 40$ & 8,963 (16.2) & 21,547 (44.3) \\
& Missing & 1,068 (1.9) & 0 (0.0) \\
\hline
\textbf{Lung cancer} & No & 48,415 (87.8) & 37,302 (76.8) \\
\textbf{family history} & Yes & 6,323 (11.5) & 10,598 (21.8) \\
& Missing & 423 (0.8) & 695 (1.4) \\
\hline
 & Underweight ($\leq 18.4$) & 295 (0.5) & 347 (0.7) \\
& Healthy weight (18.5-24.9) & 17,556 (31.8) & 13,404 (27.6) \\
\textbf{Body Mass Index}  & Overweight (25-29.9) & 23,920 (43.4) & 20,894 (43.0) \\
 (BMI) & Obesity ($\geq 30$) & 12,631 (22.9) & 13,696 (28.2) \\
& Missing & 759 (1.4) & 234 (0.5) \\
\hline
\textbf{Lung cancer} & Negative & 52,409 (95.0) & 47,084 (96.9) \\
\textbf{diagnostic} & Positive & 2,752 (5.0) & 1,511 (3.1) \\
\hline
\end{tabular}
\caption{Characteristics and statistics for PLCO and NLST groups.}
\label{tab:feature_descr}
\end{table}

\end{document}